\documentclass[lettersize,journal]{IEEEtran}
\usepackage{amsmath,amsfonts}
\usepackage{algorithmic}
\usepackage{array}
\usepackage[caption=false,font=normalsize,labelfont=sf,textfont=sf]{subfig}
\usepackage{textcomp}
\usepackage{stfloats}
\usepackage{url}
\usepackage{verbatim}
\usepackage{graphicx}
\def\BibTeX{{\rm B\kern-.05em{\sc i\kern-.025em b}\kern-.08em
    T\kern-.1667em\lower.7ex\hbox{E}\kern-.125emX}}
\usepackage{balance}
\usepackage{multirow}
\usepackage{xspace}
\makeatletter
\DeclareRobustCommand\onedot{\futurelet\@let@token\@onedot}
\def\@onedot{\ifx\@let@token.\else.\null\fi\xspace}
\def\eg{\emph{e.g}\onedot} 
\def\ie{\emph{i.e}\onedot} 
 
\def\etc{\emph{etc}\onedot} \def\vs{\emph{vs}\onedot}
\def\wrt{w.r.t\onedot} 
\def\etal{\emph{et al}\onedot}

\usepackage{color}
\definecolor{dark_red}{rgb}{0.5, 0, 0}
\definecolor{red}{rgb}{.9,0,0}
\newcommand{\mj}[1]{{\color{black}#1}}

\begin{document}
\title{GCNet: Probing Self-Similarity Learning  for Generalized Counting Network}

\author{Mingjie~Wang,~
	Yande~Li,~
	Jun~Zhou,~
	Graham~W.~Taylor,~
	and~Minglun~Gong
	\thanks{Mingjie Wang is with the School
		of Science, Zhejiang Sci-Tech University,
		China. (E-mail: mingjiew@zstu.edu.cn)}
	\thanks{Yande~Li is with Lanzhou University, China, E-mail: (yande@lzu.edu.cn) and Jun~Zhou is with Dalian Maritime University, China. (E-mail: jun90@dlmu.edu.cn)}
	\thanks{Graham~W.~Taylor is with the Shool of Engineering, University of Guelph, Canada, E-mail: (gwtaylor@uoguelph.ca) and Minglun~Gong is with the School of Computer Science, University of Guelph, Canada. (E-mail: minglun@uoguelph.ca) (Corresponding Author)}}
\markboth{Journal of \LaTeX\ Class Files,~Vol.~18, No.~9, September~2020}%
{How to Use the IEEEtran \LaTeX \ Templates}

\maketitle

\begin{abstract}
The class-agnostic counting (CAC) problem has caught increasing attention recently due to its wide societal applications and arduous challenges. To count objects of different categories, existing approaches rely on user-provided exemplars, which is hard-to-obtain and limits their generality.  In this paper, we aim to empower the framework to recognize adaptive exemplars within the whole images. A zero-shot Generalized Counting Network (GCNet) is developed, which uses a pseudo-Siamese structure to automatically and effectively learn pseudo exemplar clues from inherent repetition patterns. 
In addition, a weakly-supervised scheme is presented to reduce the burden of laborious density maps required by all contemporary CAC models, allowing GCNet to be trained using count-level supervisory signals in an end-to-end manner. Without providing any spatial location hints, GCNet is capable of adaptively capturing them through a carefully-designed self-similarity learning strategy. Extensive experiments and ablation studies on the prevailing benchmark FSC147 for zero-shot CAC demonstrate the superiority of our GCNet. It performs on par with existing exemplar-dependent methods and shows stunning cross-dataset generality on crowd-specific datasets, \eg ShanghaiTech Part A, Part B and UCF\_QNRF.
\end{abstract}

\begin{IEEEkeywords}
class-agnostic counting, exemplar-free scheme, zero-shot, self-similarity learning.
\end{IEEEkeywords}

\section{Introduction}

\begin{figure}[h]
	\begin{center}
		\includegraphics[width=\linewidth]{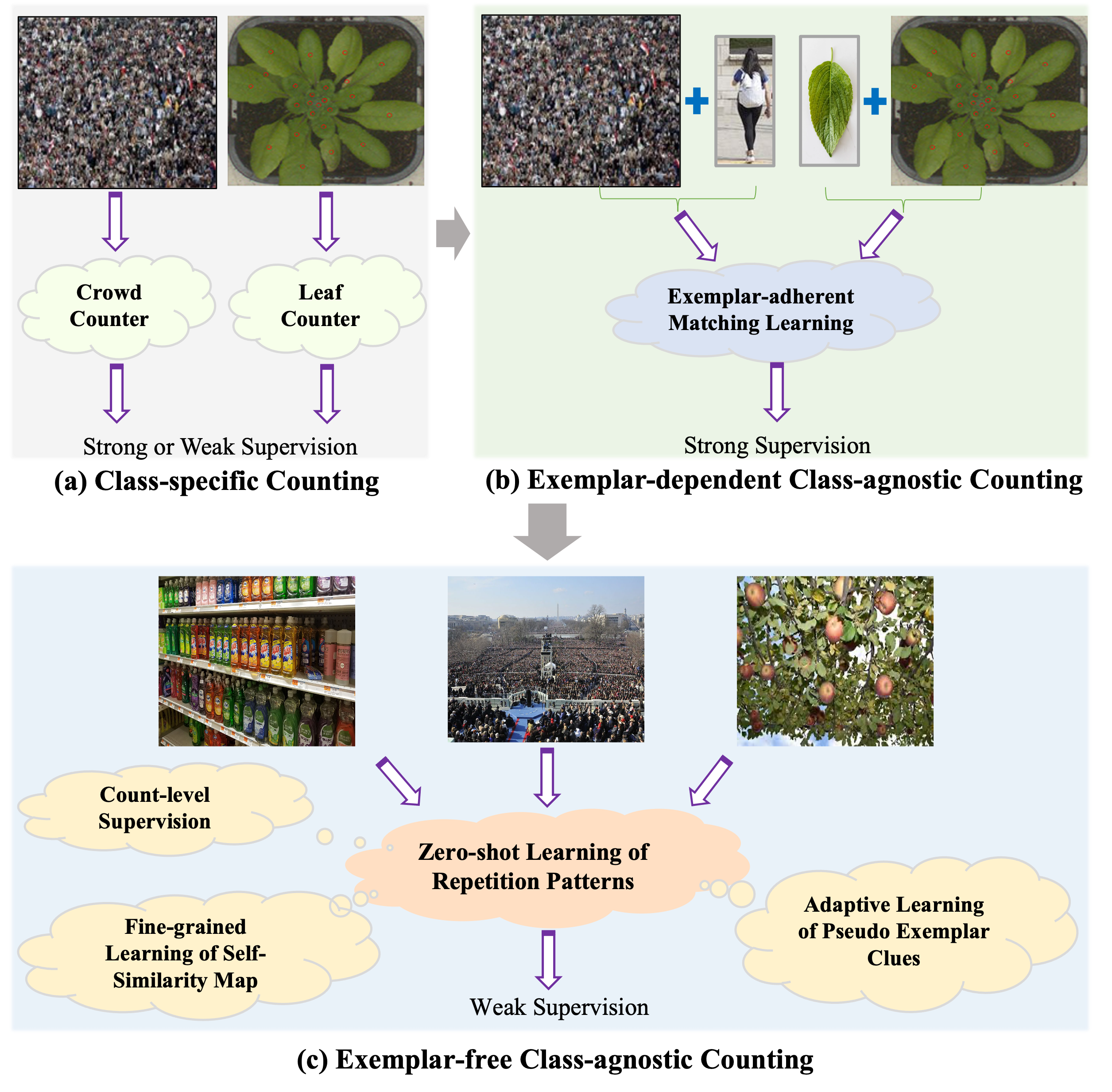}
	\end{center} 
	\caption{The evolution of counting frameworks: (a) Class-specific methods~\cite{song2021rethinking,rahnemoonfar2017deep} count objects from fixed categories; (b) Inchoate exemplar-dependent class-agnostic counters learn to match user-provided exemplars with query scenes~\cite{chattopadhyay2017counting,lu2018class,yang2021class,ranjan2021learning,shi2022represent}; (c) Our GCNet removes the requirement of exemplar annotations and learns self-similarity from repetitive object patterns. ``Strong'' supervision means laborious location-wise annotation while ``Weak'' indicates the single count-level label.}
	\label{fig:motivation}
\end{figure}

\IEEEPARstart{M}{Mainstream} Class-Specific Counting (CSC) models are conditioned to count objects of a single category, such as people~\cite{song2021rethinking}, vehicles~\cite{bui2020vehicle}, cells~\cite{khan2016deep}, and fruits~\cite{rahnemoonfar2017deep}; see Fig.~\ref{fig:motivation}~(a). Recently, the Class-Agnostic Counting (CAC) task has caught increasing attention in computer vision due to its wider application scenarios and more general nature.

In general, humans, even children, are good at subitizing. That is, estimating the size of a set without counting one-by-one. They achieve this by recognizing inherent visual repetition patterns without acquiring priors on object categories~\cite{ranjan2021learning,shi2022represent}.
To mimic humans' natural counting ability, a cohort of CAC frameworks~\cite{chattopadhyay2017counting,lu2018class,yang2021class,ranjan2021learning,shi2022represent} have been recently proposed, ushering in a promising era in the realm of object counting. They aim at subitizing or counting repetitive object instances dispersed across input scenes, thereby allowing for counting visual objects in 
broad target scenarios, \eg public spaces, traffic, agriculture, or medicine. Apart from the widespread and fascinating application targets, CAC also reduces the reliance on larger amounts of training data~\cite{shi2022represent} thanks to its few-shot learning paradigm. Stimulated by the roaring success of few-shot/meta-learning image classification~\cite{snell2017prototypical,finn2017model,peng2019few,dhillon2019baseline}, existing CAC has taken a detour from conventional fully-supervised counting approaches, and casts the object counting problem as a few-shot regression task by evaluating models on novel/unseen object categories.

Despite the persevering efforts~\cite{lu2018class,yang2021class,ranjan2021learning,shi2022represent} to enhance the learning capability of class-agnostic repetition patterns, these prominent CAC methods adopt the \emph{exemplar-matching-query} protocol that explicitly characterizes input exemplars and then matches them with the query image; see Fig.~\ref{fig:motivation}~(b). It is superfluous and even intractable to collect exemplar information (\eg bounding boxes or instance centroids) in the case of plug-and-play scenarios. Involving \emph{exemplar} regions is also paradoxical with the ultimate objective of CAC, \ie object category cues are implicitly absorbed by feeding heuristically-predefined exemplar sub-regions with single counting targets, thereby hindering the generalization of CAC to novel categories. For example, contemporary BMNet~\cite{shi2022represent} and FamNet~\cite{ranjan2021learning} both demand three manually-labelled exemplars (rectangular image patches), and fail in their absence. Apart from the dilemma of \emph{providing} exemplars, another barrier is that the performance of subsequent matching behaviour in contemporary CAC approaches is bounded by the quality/diversity of exemplar annotations, \eg shapes, scales or views.  \mj{More recently, RepRPN-Counter~\cite{ranjan2022exemplar} attempts to automatically detect several proposals/exemplars for objects in the image by introducing a two-stage  training strategy. Albeit the elimination of user-provided exemplars during the inference phase, the demand of explicitly providing exemplar labels still exists at training stage, which fails to achieve a real sense of  free exemplar. Besides, the two-stage training procedure in RepRPN is complicated and prone to limit its flexibility, thereby generating unsatisfactory results. Therefore, there is large room for improvements on effectiveness and performance of the exemplar-free CAC.}

\mj{To attenuate aforementioned issues,} in this paper, we present GCNet, a zero-shot counting framework that does not require user-provided exemplar labels \mj{during both training and inference phases while adopting one-stage end-to-end training protocol}; see Fig.~\ref{fig:motivation}~(c). It discovers exemplars by exploiting self-similarity of inherent repetition patterns.
Our goal is to probe a fully generic counting framework driven by a model with the capacity to adaptively recognize the inherent repetition patterns for everyday objects from arbitrary input images. The proposed GCNet automatically and effectively infers pseudo exemplar clues from a pseudo exemplar simulator. It further compares the extracted exemplar clues against the low-level features of the raw image to generate a high-fidelity self-similarity map as an intermediate product for the subsequent count regression. \mj{On top of the extracted exemplar hints and low-level features of the input image, our model proceeds to characterize self similarity between pseudo exemplar and raw image. This design lays a solid foundation for the generation of intermediate high-fidelity self-similarity maps.}


Current CAC algorithms also demand ground-truth location information in the form of dot/density maps as supervisory signal to guide the learning of similarity maps, which lacks self-adaptation ability and degrades their generality. Acquiring location labels is time-consuming and labour-intensive, which is usually implemented by manually placing dots in sequence on the centroids of all object instances dispersed across the whole image. In contrast, collecting weaker supervisory signals of single total count values is much easier in practical applications, \eg the real-time quantities of multifarious goods can be easily attained via a cashier system in a supermarket. Moreover, the essential purpose of CAC is to derive the total numbers of instances from natural scenes without precisely locating them, which is more imitative of human subitizing behaviour. Inspired by progress in count-level crowd counting~\cite{liang2021transcrowd,yang2020weakly,lei2021towards}, a weakly-supervised class-agnostic counting protocol is devised in our GCNet, which removes the reliance on confined/time-consuming location cues and prevents the model from trivial outcomes.

In a nutshell, the contributions in this paper are fourfold:
\begin{itemize}
	\item  \textbf{In use:} A pseudo exemplar simulator is developed to automatically and effectively learn pseudo exemplar cues. From this, a self-similarity learning scheme is designed to capture a semantically-aware similarity map. \mj{It marries our method with the fully exemplar-free property during both training and inference phases while avoiding the burdensome two-stage training.} 
	\item \textbf{Data collection:} Weakly-supervised regression eliminates the need for density maps obtained by labour-intensive point labels. This allows GCNet to be trained end-to-end using only a count-level supervisory signal. Together with the exemplar-free property, this makes GCNet require much less annotator effort.
	\item \textbf{Performance:} \mj{The state-of-the-art accuracy is achieved on the prevailing benchmark FSC147~\cite{ranjan2021learning} compared to the exemplar-free RepRPN-Counter and the BMNet-based baseline.} Impressive results on FSC147 also demonstrate the superiority of our GCNet compared to traditional CAC approaches using both exemplars and location annotations.
	\item \textbf{Application scope:} Repurposing our GCNet for challenging crowd-specific datasets (\eg ShanghaiTech~\cite{zhang2016single}  Part A, Part B and UCF\_QNRF~\cite{idrees2018composition}) further illustrates its strong generality.
	
\end{itemize}

\section{Related Work}

\begin{figure*}[t]
	\begin{center}
		\includegraphics[width=\linewidth]{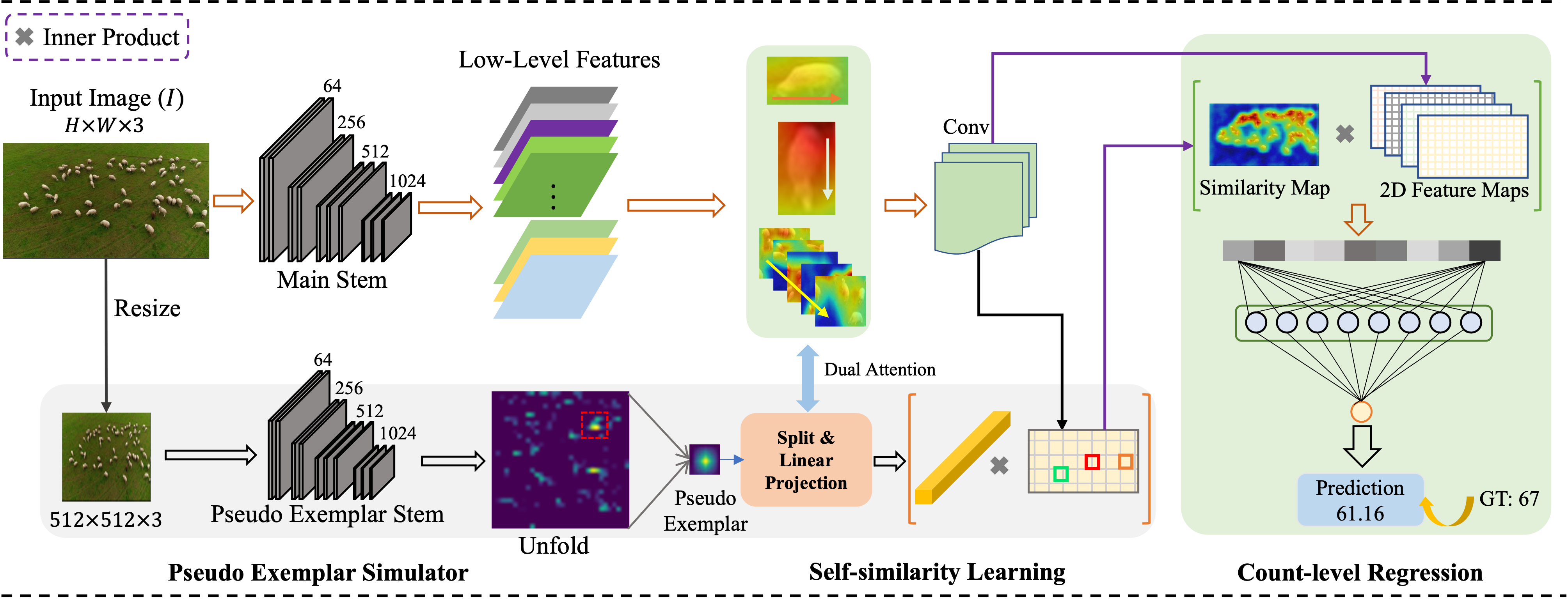}
	\end{center} 
	\caption{The overall architecture of GCNet, which features self-adaptive learning of pesude exemplar clues and weakly (count-level)-supervised training.     The delicately-designed pseudo-siamese and self-similarity learning schemes enable the generation of spatial-aware 2D similarity maps even without any location-level labels used for training.}
	\label{fig:mainstructure}
\end{figure*}

On the grounds of the generalization capability to unseen/novel class categories, existing counting methods can be roughly assigned to the following two categories.

\subsection{Class-specific Counting}
\emph{Class-Specific Counting} (CSC) models account for a larger portion of existing counting approaches. CSC aims at learning mapping relations between input scenes and target object counts/density maps on the basis of large amounts of datasets for various classes. The most-studied crowd counting methods, \eg MCNN~\cite{zhang2016single}, Hydra-CNN~\cite{onoro2016towards}, Switch-CNN~\cite{babu2017switching},
SANet~\cite{cao2018scale}, CSRNet~\cite{li2018csrnet}, ADCrowdNet~\cite{liu2019adcrowdnet}, DSSINet~\cite{liu2019crowd}, ASNet~\cite{jiang2020attention}, MBTTBF~\cite{sindagi2019multi}, Adaptive Dilated Network~\cite{bai2020adaptive}, DensityCNN~\cite{jiang2020density} and SASNet~\cite{song2021choose}, formulate the counting problem as the task of density map regression, which learns to infer pixel-level density maps. More recently, to reduce the reliance on the time-consuming annotations of pixel-wise dot maps, several pioneering studies~\cite{lei2021towards,yang2020weakly,liang2021transcrowd,wang2022crowdmlp} attempt to directly learn from weak count-level labels and further enhance the generality of crowd counting models. Towards a more generic counting framework, we are interested in directly learning from weak supervision of single count values, eliminating the need for cumbersome density maps. Apart from the popularity of crowd counting, a number of counting models for other specific categories have also been explored, such as vehicles~\cite{mundhenk2016large,hsieh2017drone,bui2020vehicle}, cells~\cite{arteta2016detecting,khan2016deep,xie2018microscopy}, fruits~\cite{rahnemoonfar2017deep,zabawa2019detection}, animals~\cite{arteta2016counting,norouzzadeh2018automatically,rivas2018detection} and bugs~\cite{ding2016automatic,zhong2018vision,wang2022recognition}.

\subsection{Class-agnostic Counting}
Conventional CSC frameworks usually require abundant annotated instances for specific types of object categories and are incapable of adapting to multifarious objects. Recently, \emph{Class-Agnostic Counting} (CAC)~\cite{chattopadhyay2017counting,lu2018class,yang2021class,ranjan2021learning,shi2022represent}  has caught increasing attention from the research community. It is focused on counting generic objects. Chattopadhyay \etal~\cite{chattopadhyay2017counting} propose to count instances of everyday objects in a patch-based manner on the COCO dataset~\cite{lin2014microsoft} with many categories. However, this approach cannot handle scenes with dense instances  as COCO is designed for object detection and the number of objects is often small. Inspired by the more than decade-old work on similarity learning~\cite{shechtman2007matching}, the \emph{exemplar-matching-query} protocol is the prevailing direction of CAC studies. Wherein, Lu \etal~\cite{lu2018class} \etal present a Generic Matching
Network (GMN) by mining the similarity between the provided exemplar patch and a query image. However, the GMN needs to be explicitly adapted to the target class categories of interest through a adaptation module. Without adaptation, GMN cannot generalize very well to novel classes~\cite{ranjan2021learning}. 

Motivated by the success of few-shot image classification ~\cite{snell2017prototypical,finn2017model,peng2019few,dhillon2019baseline} and object detection~\cite{kang2019few,fan2020few}, several related works~\cite{yang2021class,ranjan2021learning,shi2022represent} aim to mimic generalizability of humans by exploiting the few-shot counting problems. Specifically, both CFOCNet~\cite{yang2021class} and FAMNet~\cite{ranjan2021learning} leverage the spirit of Siamese to extract features for further similarity modelling. CFOCNet~\cite{yang2021class} treats the feature maps of the input exemplar as a kernel to convolve the feature maps from an input query image, whereas FAMNet~\cite{ranjan2021learning} requires explicit test-time adaptation to perform well on novel classes. It is worth noting that, to fill the void of zero-shot counting datasets, Ranjan \etal~\cite{ranjan2021learning} propose the first dataset, namely FSC147,  for training or evaluating specialized CAC models instead of using the limited COCO. \mj{More recently,}
BMNet~\cite{shi2022represent} focuses on building high-fidelity similarity maps via an explicit location supervision on intermediate similarity maps and annotated exemplar triplets. The above exemplar-dependent approaches strive to enhance the performance of CAC. However, they resort to using many extra types of annotations (\eg exemplar bounding boxes and location-wise density maps) which may not be provided or acquired in practical scenarios. In addition, they neglect the enlarging of receptive fields of exemplar tokens and the adaptation of object shapes, thereby reducing the generalization of these CAC models. \mj{The most related to our GCNet is RepRPN-Counter~\cite{ranjan2022exemplar} which firstly trains the model to identify exemplars from repeating instances in an input scene, and then performs BMNet-like \emph{exemplar-matching-query} operation. Although RepRPN has made some attempts to get rid of user-provided exemplar annotations, the requirement of exemplar labels is still needed during the training phase and the two-stage training strategy tends to hinder the generalization ability of the model, thereby producing unsatisfactory results. To capture underlying self-similarity pattern, we propose a pseudo exemplar simulator to mine exemplar hints in a one-stage and effective manner.}

\section{Exemplar-free GCNet}
Our GCNet aims to make few-shot CAC methods more general and useful through eliminating the reliance on 1) hard-to-obtain or unavailable exemplar annotations during training or inference phases and 2) labour-intensive pixel-wise supervisory signals. The overall schema of our GCNet is depicted in Fig.~\ref{fig:mainstructure}. In this section, we elaborate on each of the three pivotal components. 

\subsection{Pseudo Exemplar Simulator}

\begin{figure}[t]
	\begin{center}
		\includegraphics[width=\linewidth]{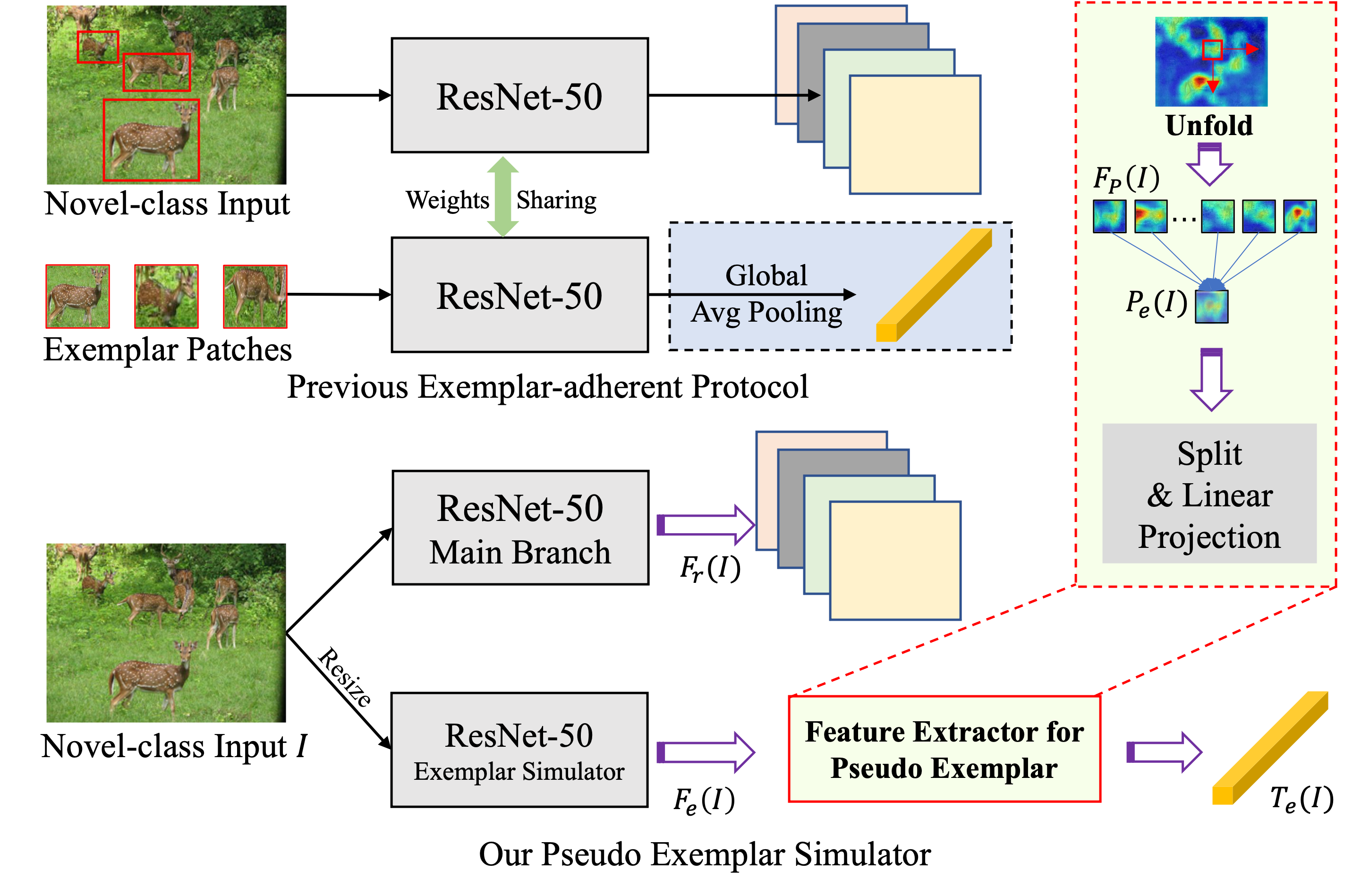}
	\end{center} 
	\caption{The comparison between the previous exemplar-dependent protocol (top) and our pseudo exemplar simulator (bottom). Our simulator recovers exemplar information of repetition patterns without any user-provided annotations.} 
	\label{fig:selfexemplar}
\end{figure}

The core motivation  of the GCNet is to remove the need for exemplar annotations from existing CAC and realize the true sense of generic counting. To this end, a pseudo-Siamese simulator is constructed to capture pseudo exemplar clues from the resized raw image without being explicitly fed exemplar sub-regions. As demonstrated in Fig.~\ref{fig:selfexemplar}, differing from the inchoate exemplar-based approaches built upon weight-sharing 
Siamese structures, our two branches are separately optimized to extract discriminative representations more suitable for the learning of adaptive exemplar cues. Specifically, given a natural scene $I \in R^{H\times W\times 3}$ of arbitrary size, the main branch extracts primary 2D feature maps with channel dimension $C$, denoted as $F_r(I) \in R^{\frac{H}{16}\times\frac{W}{16}\times C}$. The secondary branch resizes $I$ to $512\times 512$ to normalize instances with varied shapes/scales and to reduce computational overhead. It then generates exemplar-focused outputs $F_e(I) \in R^{32\times 32 \times C}$. 

\paragraph{Extracting a Pseudo Exemplar Patch.} 
Motivated to achieve higher efficiency, we propose to operate on the high-level feature outputs $F_e(I) \in R^{32\times 32 \times C}$ of our simulator instead of traversing the raw input image. Analogous to the implementation of convolution, we use the ``\emph{unfold}'' operation $U(\cdot)$ with kernel size of 8$\times$8 and stride of 1 to cover a wide range of instance regions. Then a sequence of feature patches are unfolded, $F_P(I) = U(F_e(I)) = \{F^p_{1}, F^p_{2}, F^p_{3}, ... F^p_{n}\}, F^p_{i} \in R^{8\times 8 \times C}$ where $i$ represents the $i^{th}$ unfolded patch. To collapse the extra dimension of patch number and acquire more common attributes of pseudo exemplar, we impose a pixel-level average operation along the patch dimension and produce the pseudo exemplar patch $P_e(I)=\frac{1}{n}(F^p_{1} + F^p_{2} + F^p_{3} + ... + F^p_{n}),$ where $P_e(I) \in R^{8\times 8 \times C}$. Inspired by the success of vision transformers~\cite{dosovitskiy2020image,liu2021swin} on enlarging receptive fields, we split $P_e(I)$ into 16 feature sub-patches of size 2$\times$2. All patches are flattened to a 1D token with the shape of $(B,16,C)$ where $B$ is the batch size, and then are transformed through linear projection with dimension of $C$. Finally, the proposed simulator outputs the crude 2D feature maps $F_r(I) \in R^{\frac{H}{16}\times\frac{W}{16}\times C}$ and the pseudo exemplar 1D token $T_e(I) \in R^{16C}$.

\subsection{Dual-Attention Self-Similarity Learning}
\begin{figure}[h]
 	\begin{center}
 		\includegraphics[width=\linewidth]{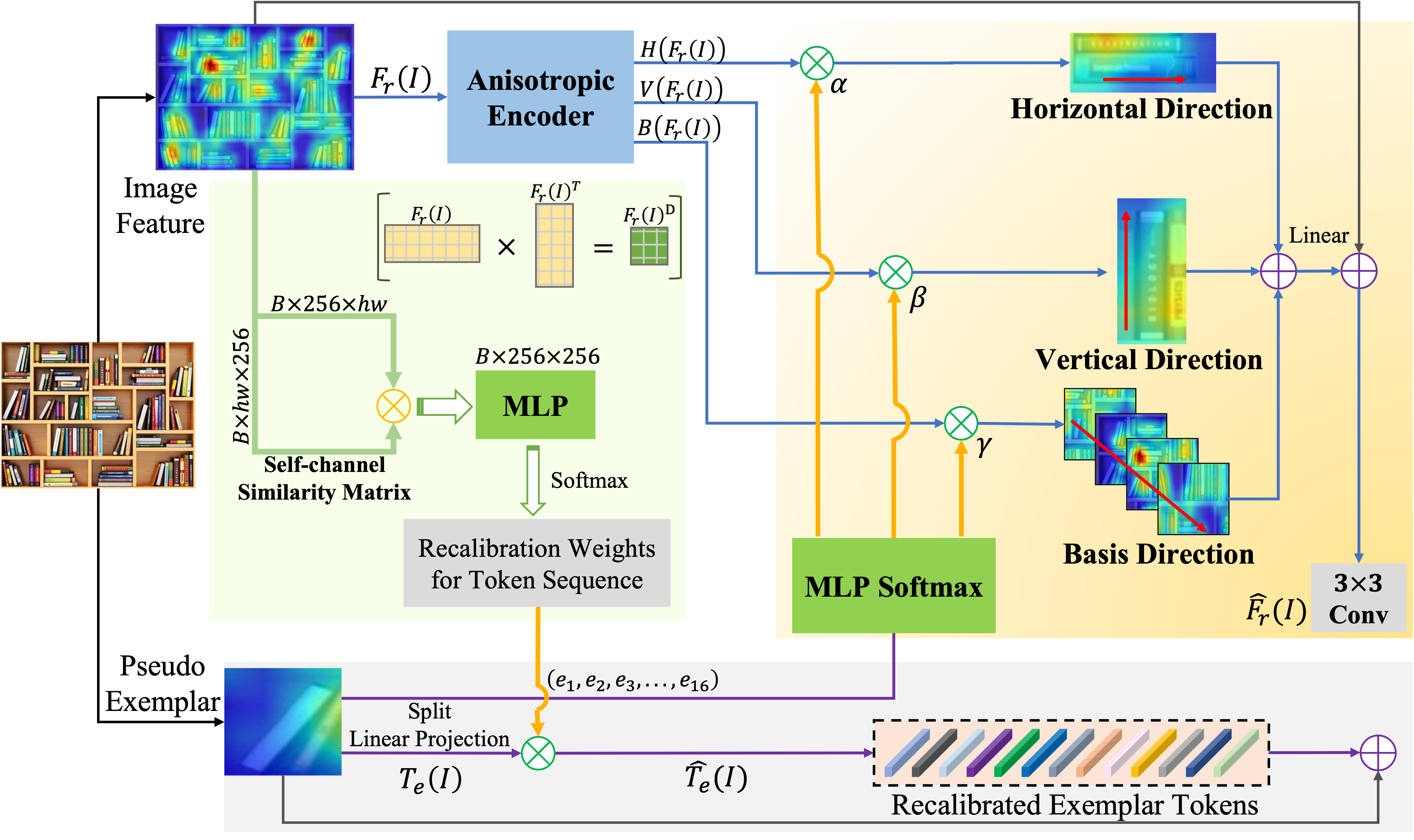}
 	\end{center} 
 	\caption{The pipeline of our dual-attention self-similarity learning, which includes an anisotropic encoder and a dual-attention condenser.} 
 	\label{fig:similarity}
 \end{figure}

In the next step, the extracted $F_r(I)$ and $T_e(I)$ are leveraged to construct a high-quality self-similarity map. A \emph{dual-attention self-similarity} (DASS) learning protocol is designed for this task; see Fig.~\ref{fig:similarity}.

\paragraph{Anisotropic Encoder.}
In practical scenarios, object instances within images in-the-wild often vary in shapes or orientations, which poses great challenges to fine-grained feature learning and similarity awareness. For example, as shown in Fig.~\ref{fig:similarity}, books can be placed vertically or horizontally. In this circumstance, separate convolutions with kernels along different directions (\eg the horizontal or vertical axes) enables the capture of discriminative features covering a larger range of shapes. To this end, encouraged by the resounding success of anisotropic filtering in image processing~\cite{gerig1992nonlinear}, we propose a novel anisotropic encoder in our DASS to probe anisotropic characteristics along three directions in feature space, \ie horizontal, vertical and basis/channel directions. 
Specifically, given the input spatial representations $F_r(I)$, three kinds of kernels (orienting horizontal, vertical and channel dimensions) are leveraged to convolve $F_r(I)$, computing anisotropic features $H(F_r(I))$, $V(F_r(I))$ and $B(F_r(I))$. By doing so, features in each dimension are disentangled from the conventional isotropic convolutions used in contemporary CAC models. 

\paragraph{Dual-attention Condenser.}
Even though the pseudo exemplar token $T_e(I)$ and anisotropic features $H(F_r(I))$, $V(F_r(I))$ and $B(F_r(I))$ have rich feature representations, the interaction/communication between two sources are ignored, which is detrimental to the matching quality. To further refine these features, a dual-attention condenser is posed here to distil knowledge across the branches. As shown in Fig.~\ref{fig:similarity}, $T_e(I)$ is passed to a MLP-based unit with softmax activation. Then, three normalized soft weights, $\alpha$, $\beta$ and $\gamma$, are computed and fed into the anisotropic encoder to emphasize the most informative directions in feature space, while suppressing useless ones. Therefore, we can obtain the integrated anisotropic features as follows:
\begin{equation}
\bar{F}_r(I)=W\ast(F_r(I) + \alpha H(F_r(I)) + \beta V(F_r(I)) + \gamma B(F_r(I))),
\label{equ:attention1}
\end{equation}
where $\alpha+\beta+\gamma=1$ and $+$ denotes pixel-wise summation, whereas $\ast$ indicates 3$\times$3 convolution and $W$ is a matrix of trainable filters.

Accompanying the soft recalibration imposed on the anisotropic features, a right-about attention mechanism is also designed to steer the learning of pseudo exemplar cues under the guidance of knowledge distilled from the main branch. In detail, to model the similarity among self-channel filters, the matrix product between feature maps $F_r(I)  \in R^{256\times hw}$ and its transpose $F_r(I)^{\top}\in R^{hw\times 256}$ is carried out to produce a square matrix  $F_r(I)^D \in R^{256\times 256} = F_r(I)F_r(I)^{\top}$, namely a self-channel similarity matrix. By doing so, we also enable the module to cater to arbitrary resolutions. Through the similarity matrix among feature maps of channel filters, the implicit repetitive patterns can be tracked. To fully make use of the similarity among the set of channel filters, we utilize  $F_r(I)^D$ to generate attention weights $(e_1, e_2, e_3, \dots, e_{16})$ for guiding the learning of discriminative exemplar tokens. We formulate this recalibration as
\begin{equation}
\bar{T}_e(I)= [e_1T^e_{1}(I), e_2T^e_{2}(I) , \dots,  e_{16}T^e_{16}(I)] + T_e(I),
\label{equ:attention2}
\end{equation}
where $e_1 + e_2 + e_3 + \dots + e_{16} = 1$ and $\left[ \cdot \right]$ represents concatenation. The outputs of this module include the integrated anisotropic features $\bar{F}_r(I)$ and recalibrated exemplar token $\bar{T}_e(I)$. They are then further aggregated to acquire the self-similarity maps. Following common practice in exemplar-dependent approaches~\cite{shi2022represent,ranjan2021learning}, the final self similarity matrix  $S(I) \in R^{h\times w}$ with regions of interest can be calculated by $S(I) = \frac{1}{16}(\bar{F}_r(I)\bar{T}_e(I)^{\top})$.

\subsection{Weakly-supervised Location-aware Counter}

\begin{figure}[h]
	\begin{center}
		\includegraphics[width=\linewidth]{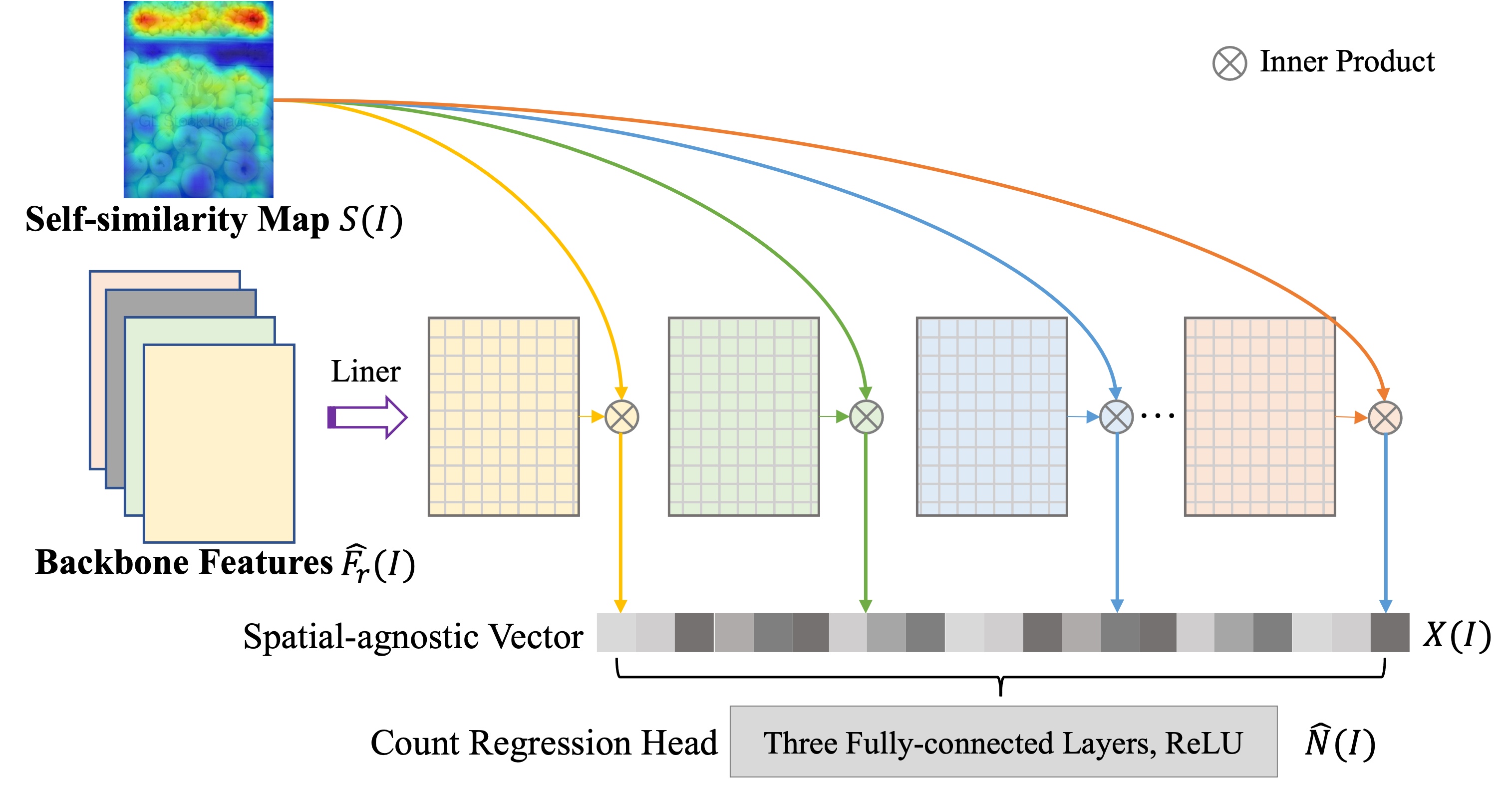}
	\end{center} 
	\caption{The design scheme of the proposed weakly-supervised location-aware counter. It combines the self-similarity map with the low-level feature map to exploit global correlations over spatial pixels while reducing dimensionality to produce a scalar count.}
	\label{fig:counter}
\end{figure}

\paragraph{Location-aware Counter.} Having recourse to the extracted/learned  intermediate elements, \eg crude image features $F_r(I)$, pseudo exemplar token $T_e(I)$, integrated anisotropic features $\bar{F}_r(I)$, recalibrated exemplar token $\bar{T}_e(I)$, and the self-similarity map $S(I)$, our GCNet enters the final phase that proceeds to predict the number of object instances over the input scene. To fully exploit the cues within the generated similarity map while boosting the correlation between similarity map and feature maps of the raw image, we devise a  weakly-supervised yet location-aware counter (see Fig.~\ref{fig:counter}) to directly estimate count numbers for arbitrary input scenes. Specifically, the correlation between $\bar{F}_r(I)$ and $S(I)$ is computed by the inner product along the channel dimension, and then a high-level spatial-agnostic vector $X(I)$ is produced.
\begin{equation}
X(I) = \bar{F}_r(I)\cdot S(I), X(I) \in R^{1\times C},
\label{equ:counter}
\end{equation}
where $\cdot$ denotes the inner product and $C$ is the feature dimension. Then a regression head with three linear layers is constructed to directly map the vector $X(I)$ to the final predicted count $\hat{N}(I)$ for input image $I$.

\paragraph{Unstructured Supervisory Signals.} 
To boost the generality of the CAC framework, GCNet also eliminates the reliance on pixel-level ground truths (\ie dot or density maps) by utilizing only integer count values as supervisory signals. Through the inner products between the similarity map and feature maps, the removal of 2D structure hints at this stage is conducive to the modelling of high-level repetition patterns while allowing for improved generalization to scenes at arbitrary resolutions. Given the spatial-agnostic vector $X(I)$, three fully-connected layers with widths of 64, 32 and 1 respectively are configured to predict the final instance count $\hat{N}(I)$. Instead of using the heuristically-defined loss terms in the latest BMNet~\cite{shi2022represent} to constrain the similarity map and predicted density map simultaneously, we adopt a simple yet effective $L_2$ Euclidean loss as the objective function $L$ to optimize our framework without adjusting any hyperparameters:
\begin{equation}
L = \frac{1}{B}\sum_{B}^{i}||\hat{N}(i)-N(i)||_2^2,
\label{equ:lossfunction}
\end{equation}
where $B$ is the batch size and $N(i)$ indicates the ground-truth count value for the $i^{th}$ input scene.

\paragraph{The Comparison of Regression Architectures between Baseline and our GCNet.}
\mj{Here we show the architecture details of counters in the BMNet-based Baseline and our GCNet aiming for demonstrating the difference between two types of regression heads clearly.  It can be observed that our counter focuses on a set of low-cost linear projections and eliminates the reliance on fixed input image sizes trough ``BMM'' between intermediate self-similarity maps and basic 2D features.}

\begin{table}[h]
	\begin{center}
		\setlength{\tabcolsep}{7.5mm}
		\begin{tabular}{c|c}
			\hline
			\multicolumn{2}{c}{ConvNet Configurations}\\
			\hline
			BMNet-based Baseline & Our Counter in GCNet\\
			\hline
			Concatenation & Splitting \\ 
			\hline
			Conv7-196 & Linear-128\\
			Max Pooling & Linear-1\\
			\hline
			Conv5-128 & \multirow{2}{*}{BMM}\\
			Max Pooing & \\
			\hline
			Conv3-64 & \multirow{2}{*}{Linear-64}\\
			Global Avg Pooling & \\
			\hline
			Conv1-32 & Linear-32\\
			\hline
			Conv1-1 & Linear-1\\
			\hline
		\end{tabular}
	\end{center}
	\caption{The architecture difference between the BMNet-based counter in baseline and the regression head in our GCNet. Here ``BMM'' denotes a matrix-matrix product of matrices.}
	\label{table:countercomparison}
\end{table}

\section{Experiments and Analysis}

\paragraph{Datasets.} Following the state-of-the-art CAC approaches~\cite{ranjan2021learning, shi2022represent}, FSC147~\cite{ranjan2021learning} is adopted in our experiments since it is a unique large-scale dataset specifically designed for evaluating CAC. It constitutes 6,135 natural images of 147 object categories including fruits, books, skis, \etc. To verify the generality of CAC models, categories in three sub-sets (3,659 training, 1,286 testing, and 1,190 validation samples) have no overlap. Although corresponding exemplars and density maps are provided, they are not used by GCNet.
To further verify its generalization capability, GCNet is also tested on three crowd counting datasets: ShanghaiTech~\cite{zhang2016single} Part~A (182 images), Part~B (316 images), and  UCF\_QNRF~\cite{idrees2018composition} (334 images).

\paragraph{Implementation Details.} For fair comparisons, we use the same strategy to resize the original image as in FAMNet~\cite{ranjan2021learning} and BMNet~\cite{shi2022represent}. During the training phase, input images are resized within [384,1584] pixels while keeping the same aspect ratio for the main backbone. For the pseudo exemplar simulator stem, however, all images are resized to 512$\times$512 as previously discussed. Two frontend stems for extracting basic features and pseudo exemplar features are built using the first four blocks of a SwAV~\cite{caron2020unsupervised}-initialized ResNet50~\cite{he2016deep}. Random scaling, horizontal flip, vertical flip, and cutout~\cite{devries2017improved} are incorporated to augment the training sets. The batch size is set to 10 while the fixed learning rate is $1e^{-5}$. GCNet is developed on PyTorch~\cite{paszke2019pytorch} and trained in an end-to-end manner using the AdamW~\cite{loshchilov2017decoupled} optimizer.

\subsection{Comparison with State-of-the-art}

\begin{table*}[h]
	\begin{center}
		\setlength{\tabcolsep}{4.5mm}
		\begin{tabular}{ccccccc}
			\hline
			Frameworks & Exemplar & Location & {\bf MAE} for \emph{Val}  & {\bf MSE} for \emph{Val} & {\bf MAE} for \emph{Test} & {\bf MSE} for \emph{Test}\\
			\hline
			GMN~\cite{lu2018class} & \checkmark & \checkmark & 29.66 & 89.81 & 26.52 & 124.57 \\
			FamNet~\cite{lu2018class} &  \checkmark & \checkmark & 24.32 &70.94 & 22.56 & 101.54 \\
			FamNet+~\cite{lu2018class} &  \checkmark & \checkmark & 23.75 & 69.07 & 22.08 & {\bf 99.54}\\
			CFOCNet~\cite{yang2021class} &  \checkmark & \checkmark & 21.19 & {\bf 61.41} & 22.10 & 112.71\\
			BMNet~\cite{shi2022represent} &  \checkmark & \checkmark & {\bf 19.06} & 67.95 & {\bf 16.71} & 103.31\\
			\hline
			RepRPN-Counter~\cite{ranjan2022exemplar}  &  $\times$ &  $\times$ & 29.24 & 98.11 & 26.66 & 129.11 \\
			Baseline &  $\times$ &  $\times$ & 23.14 & 77.30 & 21.88 & 112.37\\
			GCNet+Exemplar (ours) & \checkmark &  $\times$ & 19.61 & 66.22 & 17.86 & 106.98 \\
			GCNet (ours) &  $\times$ &  $\times$ & {\bf 19.50} & {\bf 63.13} & {\bf 17.83} & {\bf102.89}\\
			\hline
		\end{tabular}
	\end{center}
	\caption{Quantitative comparisons with SOTA exemplar-based approaches and the baseline model on validation and test sets of FSC147. 
		The 2nd and 3rd columns show the requirements on exemplars and pixel-level density maps, respectively. The best results for exemplar-dependent and -free methods are in boldface.}
	\label{table:comparison}
\end{table*}

\begin{figure}[h]
	\begin{center}
		\includegraphics[width=\linewidth]{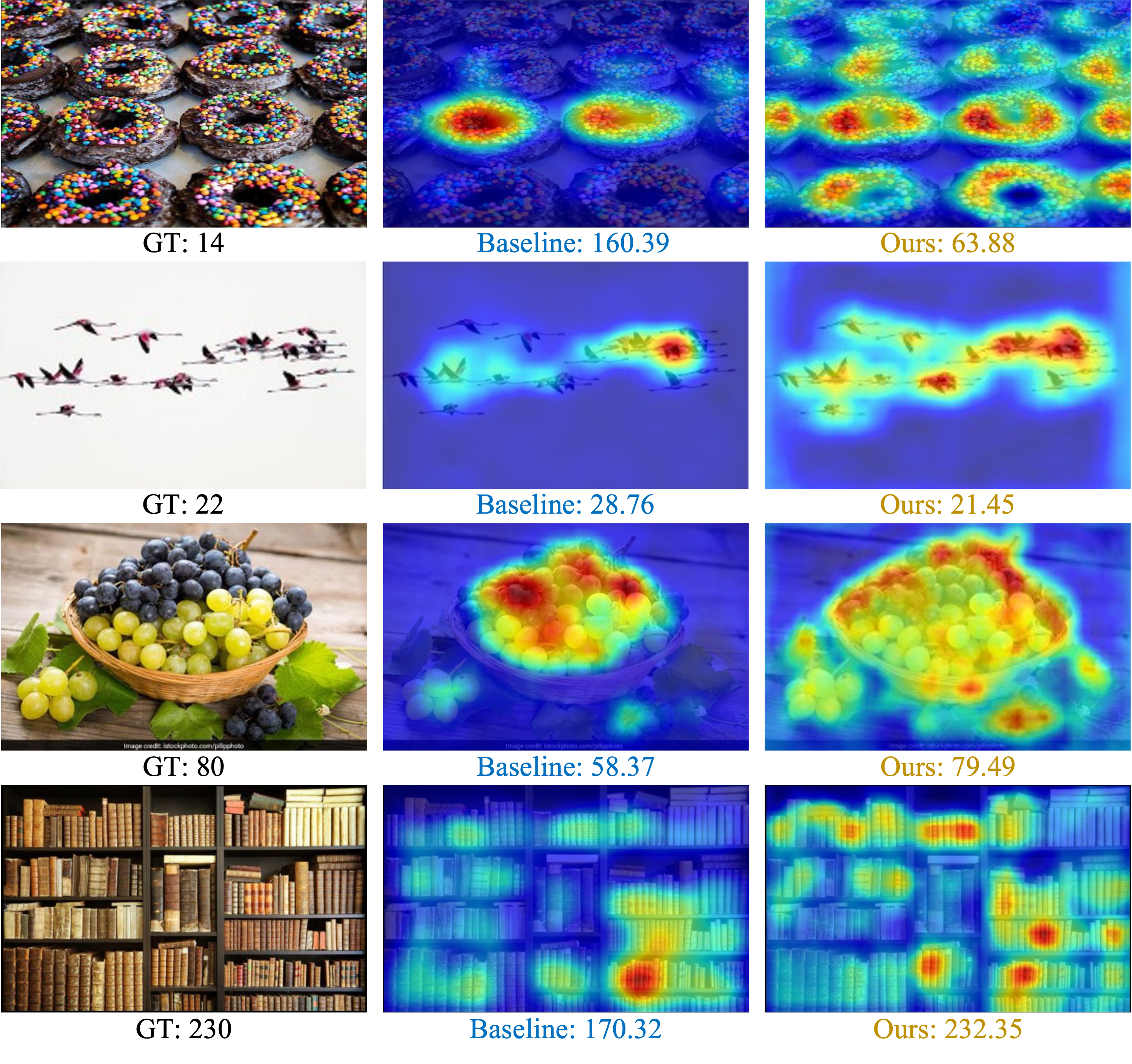}
	\end{center} 
	\caption{Visualization results on the \emph{val} set of FSC147 dataset. The examples exhibit significant challenges, \eg ambiguous counting categories (doughnuts or sprinkles on doughnuts) and extreme cardinality variation (from 14 to 458). Compared with the BMNet-based baseline, GCNet directly predicts accurate counts, while adaptively producing spatially-aware self-similarity maps.} 
	\label{fig:valexample}
\end{figure}

\paragraph{Baseline: Exemplar-fee CAC based on BMNet.}
The existing CAC approaches~\cite{ranjan2021learning, shi2022represent} all abide by the conventional exemplar-matching-query protocol to construct a similarity map between explicit exemplars and a query image. For a fairer comparison, we construct an exemplar-free baseline based on the similarity matching in SOTA BMNet~\cite{ranjan2021learning}. Specifically, we revise the backbone of BMNet by introducing an additional stem to stimulate the pseudo exemplar features, like the one in GCNet. To allow the baseline to output directly a single unconstrained value, global average pooling is plugged into the counter of BMNet while changing all upsampling operations to max pooling layers. 

\paragraph{GCNet+Exemplar: Exemplar-guided GCNet.}
To exploit the effects of exemplar annotations on our weakly-supervised framework, we also build an exemplar-guided GCNet, which attaches an additional ResNet-50 stem to the frontend of our GCNet for processing labelled exemplar patches. The exemplar features are then fed into the same back-end structure of GCNet to predict an auxiliary single count $\hat{N}_{e}$. For the GCNet+Exemplar, we update the objective function $L$ in Eq.~\ref{equ:lossfunction} as follows:
\begin{equation}
L_{e} = 0.5\frac{1}{B}\sum_{B}^{i}||\hat{N}_{e}(i)-N(i)||_2^2+0.5L.
\label{equ:loss2}
\end{equation}

\paragraph{CAC-oriented FSC147.}
We compare GCNet with prominent state-of-the-art exemplar-based approaches, \mj{the latest exemplar-free RepRPN-Counter and the BMNet-based backbone in Tab.~\ref{table:comparison}. The results show that GCNet remarkably outperforms the state-of-the-art RepRPN-Counter by 33.31\%  MAE and 34.98\% on the validation set while 33.12\% MAE and 20.30\% MSE on the test set. These significant gains demonstrate that the capability of our exemplar-free scheme surpasses the current RepRPN-Counter even only weak supervision is used for training the model (\ie no information is provided regarding instance locations), and our GCNet achieves the state-of-the-art results in the scope of exemplar-free CAC algorithms. In addition, GCNet also outperforms the designed BMNet-based} baseline by 15.73\%  MAE and 18.33\%  \emph{w.r.t} the validation set (\emph{Val}) whereas attaining improvements of 18.51\% MAE and 8.43\% MSE \emph{w.r.t} the test set (\emph{Test}). A few example images and their self-similarity maps are depicted in Fig.~\ref{fig:valexample}. Note that even the baseline obtains lower MAE values (23.14 and 21.88) on \emph{Val} and \emph{Test} compared to the conventional FamNet (24.32 and 22.56) \& FamNet+ (23.75 and 22.08)  that require laborious annotations of both exemplar and density map. This suggests that our strong baseline possesses the capacity to adaptively mine pseudo exemplar cues. It can be observed that GCNet outperforms CFOCNet and secures the second place of MAEs compared with all CAC methods, \ie they are slightly behind the SOTA BMNet (0.44$\uparrow$ MAE on \emph{Val} and  1.12$\downarrow$ MAE on \emph{Test}). Interestingly, the adoption of explicit exemplar labels into our GCNet (``GCNet+Exemplar") slightly degrades the plain GCNet to 19.61 MAE on \emph{Val} and 17.86 MAE on \emph{Test}. Our hypothesis for this phenomenon is that the additional exemplar guidance confuses the GCNet backbone on what to learn. That is, providing only selected exemplars without information on other object instances' locations (\ie dot/density maps) may encourage GCNet to only look for instances that have similar appearance/style as the exemplars.

\paragraph{Few-shot Generalization to Crowd Datasets.}
To illustrate the generality of GCNet, we also carry out cross-dataset verification on three popular benchmark datasets for the crowd counting task: ShanghaiTech Part~A~\cite{zhang2016single}, Part~B~\cite{zhang2016single} and UCF\_QNRF~\cite{idrees2018composition}. Even though the training set of FSC147 includes crowd scenes, they are extremely scarce (19 images) and with low crowd densities. The three crowd counting datasets, on the other hand, contain highly-congested scenes and have drastic scale/density variations. The comparison between GCNet and contemporary transfer-learning-based crowd counters (source domain still is crowd-specific)~\cite{gao2021domain, han2020focus, wang2019learning} is shown in Tab.~\ref{table:crossdata}. The quantitative results illustrate that, even with a sparse set of only 19 samples of people, the transferability of our few-shot GCNet outperforms the representative density map-based MCNN~\cite{zhang2016single} and transfer-learning approaches~\cite{gao2021domain, han2020focus, wang2019learning}, all of which are trained using hundreds of crowd images. The comparison results show that our GCNet attains impressive performance on par with crowd-specific models even seeing only 19 people scenes as well as without any location cues. Several visualization examples of learned self-similarity maps on three crowd datasets are depicted in the Supplementary.

\begin{table}[h]
	\begin{center}
		\setlength{\tabcolsep}{2mm}
		\begin{tabular}{ccccccc}
			\hline
			\multirow{2}{*}{Methods}& \multicolumn{2}{c}{{\bf Part~A}} 
			&\multicolumn{2}{c}{{\bf Part~B}}&\multicolumn{2}{c}{{\bf UCF\_QNRF}}\\
			\cline{2-7}
			~& MAE & MSE & MAE & MSE & MAE & MSE\\
			\hline
			MCNN~\cite{zhang2016single} & 221.4 & 357.8 &  85.2 & 142.3 & - & -\\
            NoAdpt in ~\cite{gao2021domain} & 206.7 & 297.1 & 24.8 & 34.7 & 292.6 & {\bf 450.7}\\
            NoAdpt in ~\cite{han2020focus} & 190.8 & 298.1 & 24.6 & 33.7 & 296.1 & 467.9\\
            NoAdpt in ~\cite{wang2019learning} & {\bf 160.0} & {\bf 216.5} & {\bf 22.8} & {\bf 30.6} & {\bf 275.5} & 458.5\\
            \hline
            Baseline & 256.6 & 372.8 &  63.4 & 95.4 & 532.5 & 838.5\\
			GCNet (ours)& {\bf 148.9} & {\bf 260.7} & {\bf 38.6} & {\bf 53.9} & {\bf 294.2} & {\bf 541.6}\\
			\hline
		\end{tabular}
	\end{center}
	\caption{Testing the pre-trained GCNet on almost unseen crowd scenes without any fine-tuning.}
	\label{table:crossdata}
\end{table}

\begin{figure}[h]
	\begin{center}
		\includegraphics[width=\linewidth]{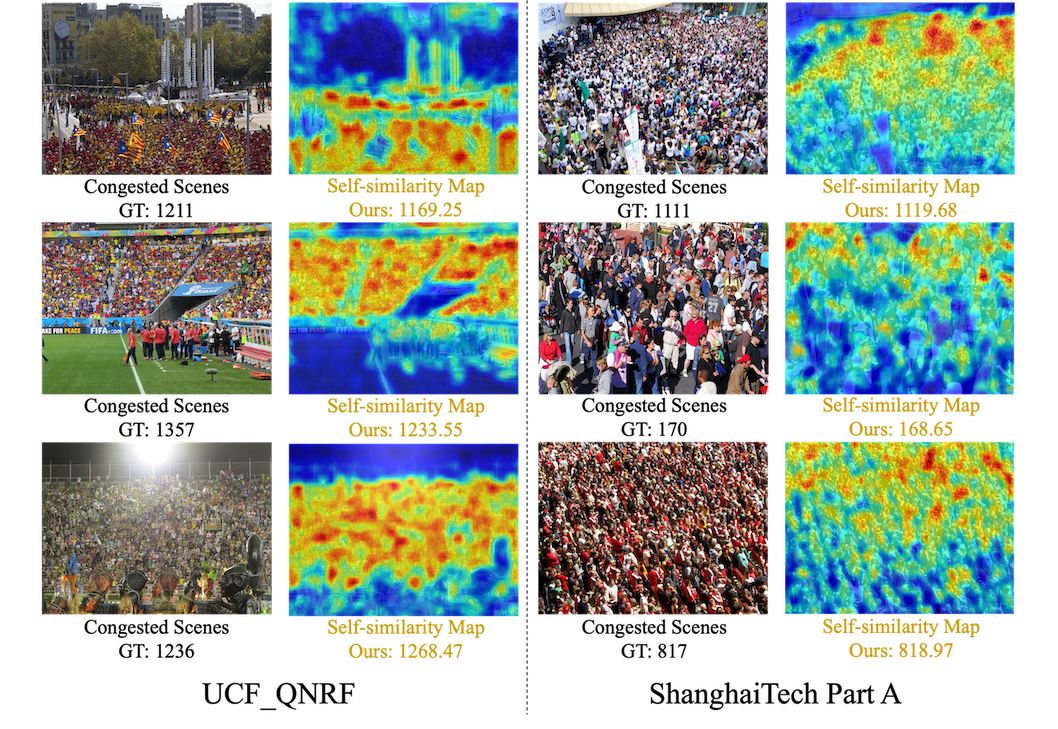}
	\end{center} 
	\caption{The visualization examples of similarity maps and prediction results provided by Baseline and our GCNet on crowd-specific UCF\_QNRF and ShanghaiTech Part A datasets, respectively, which illustrates the strong generality of our GCNet.} 
	\label{fig:crowd}
\end{figure}

To visually demonstrate the attractive generality of our GCNet, we show our predictions and intermediate self-similarity maps for several congested samples with over hundreds of people in UCF\_QNRF and ShanghaiTech Part A; see Fig.~\ref{fig:crowd}. It can be observed that, even with everyday objects (\eg apples and woods) in training set, our model achieves impressive performance on almost unseen crowd categories. In addition, compared with the Baseline, our GCNet attains attractive prediction results, which demonstrates the stunning impacts of our proposed components on the transferability of the GCNet.

\subsection{Ablation Study}
\paragraph{The Impacts of the Proposed Components.}
We perform ablation studies on the validation and test sets of FSC147 to assess the impacts of individual components. In Tab.~\ref{table:abunits} we denote with $\checkmark$ the addition of components to our baseline with pseudo exemplar simulator that include: self-similarity learning ($S$), dual-attention condenser ($D$), anisotropic features/exemplar token recalibration ($M$), and the weakly-supervised location-aware counter ($C$). We also test multiple components in the last two rows, with our final model represented in the last row.

\begin{table}[h]
	\begin{center}
		\setlength{\tabcolsep}{2.8mm}
		\begin{tabular}{cccccccc}
			\hline
			\multirow{2}{*}{No.} & \multicolumn{2}{c}{{\bf \emph{S}}} & \multirow{2}{*}{C} & \multicolumn{2}{c}{{\bf \emph{Val}}} 
			&\multicolumn{2}{c}{{\bf \emph{Test}}}\\
			\cline{2-3}
			\cline{5-8}
			~ & M & D & ~ & MAE & MSE & MAE & MSE\\
			\hline
			B0 & $\times$ & $\times$ & $\times$ & 23.14 & 77.30 & 21.88 & 112.37\\
			B1 & $\checkmark$ & $\checkmark$ & $\times$ & 23.04 & 80.08 & 21.53 & 111.43\\
			B2 & $\times$ & $\times$ & $\checkmark$ & 20.36 & 68.33 & 18.91 & 107.41 \\
			B3 & $\checkmark$ & $\times$ & $\checkmark$ & 20.03 & 66.47 & 17.99 & 107.20\\
			B4 & $\checkmark$ & $\checkmark$ & $\checkmark$ & {\bf 19.50} & {\bf 63.13} & {\bf 17.83} & {\bf102.89}\\
			\hline
		\end{tabular}
	\end{center}
	\caption{An ablation study that incrementally adds each critical component of GCNet. Results are reported on the FSC147 dataset. For each metric, the best performing entry across models is in boldface.}
	\label{table:abunits}
\end{table}

\begin{figure}[h]
	\begin{center}
		\includegraphics[width=\linewidth]{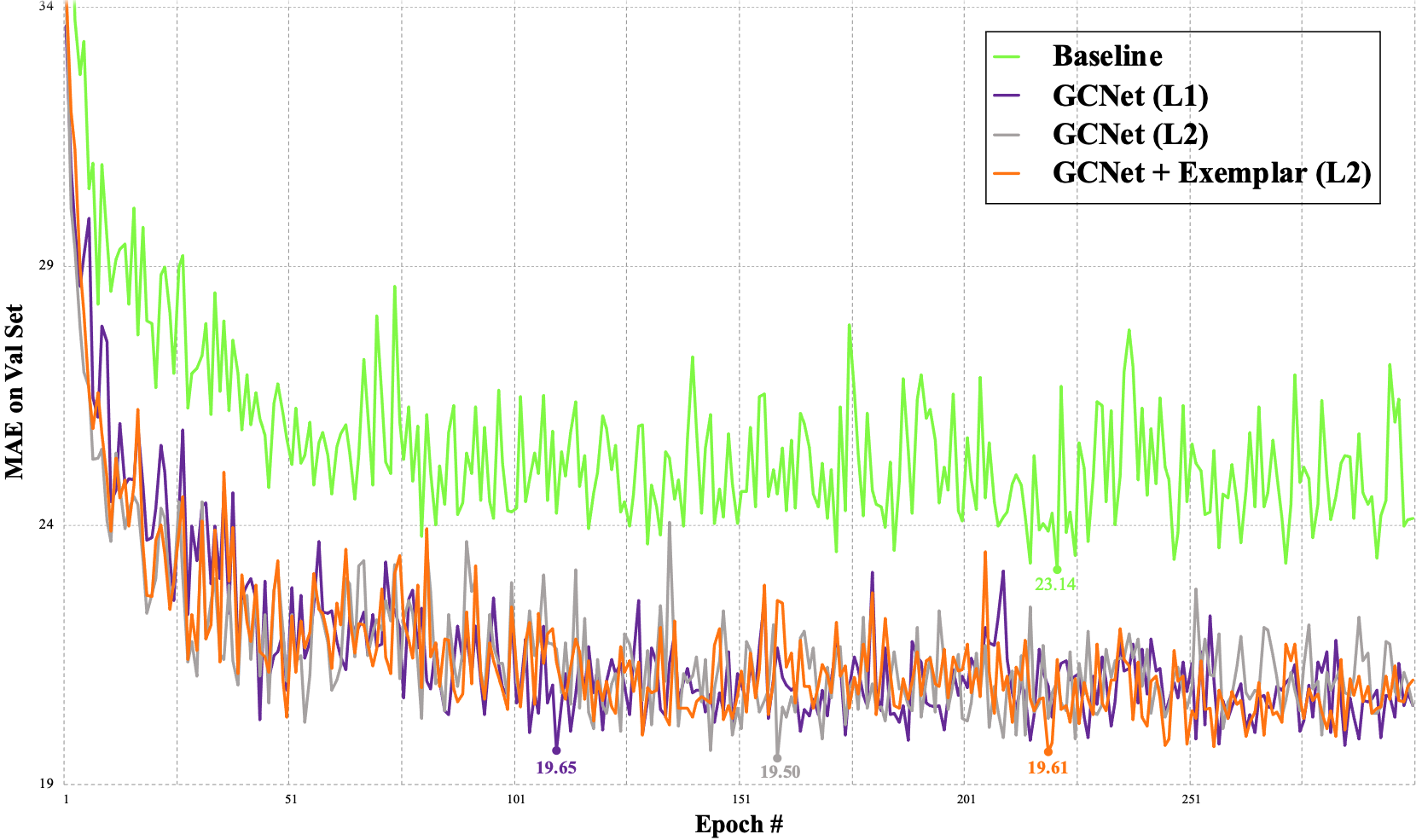}
	\end{center} 
	\caption{The curves of training loss and validation accuracy from the Baseline, GCNet with $L_1$ loss and GCNet with $L_2$ loss.} 
	\label{fig:baselinel1l2}
\end{figure}

\paragraph{$L_1$ \vs $L_2$ as Distance Metrics.}
Both $L_1$ and $L_2$ are often used as distance metrics for regression problems. To determine which yields better performance, we empirically compare the learning curves of $L_1$- and $L_2$-equipped GCNets; see orange and blue curves in Fig.~\ref{fig:baselinel1l2}. Both curves show a similar overall trend, but the $L_2$ loss delivers a slightly lower MAE (19.50 \vs 19.61) on the validation set. We also quantitatively report the results on test set in Tab.~\ref{table:abloss} provided by the original GCNet and the Exemplar-guidance GCNet (GCNet+Exemplar), which demonstrates that $L_1$ seems to yield superior generalizability of GCNet or GCNet+Exemplar on the test set than $L_2$-based models.

\begin{table}[h]
	\begin{center}
		\setlength{\tabcolsep}{2.5mm}
		\begin{tabular}{ccccc}
			\hline
			\multirow{2}{*}{Methods}& \multicolumn{2}{c}{{\bf \emph{Val}}} 
			&\multicolumn{2}{c}{{\bf \emph{Test}}}\\  
			\cline{2-5}
			~ & MAE & MSE & MAE & MSE\\
			\hline
			GCNet+Exemplar ($L_1$) & 19.61 & 67.22 & {\bf 17.00} & 105.33 \\
			GCNet+Exemplar ($L_2$) & 19.61 & 66.22 & 17.86 & 106.98 \\
			\hline
			GCNet ($L_1$) & 19.65 & 67.65 & 17.66 & 108.16 \\
			GCNet ($L_2$) & {\bf 19.50} & {\bf 63.13} & 17.83 & {\bf 102.89} \\
			\hline
		\end{tabular}
	\end{center}
	\caption{Ablation study on $L_1$ and $L_2$ distance metrics imposed on the original GCNet and GCNet+Exemplar, respectively.} 
	\label{table:abloss}
\end{table}

\subsection{Visualization and Failure Cases}

\begin{figure}[h]
	\begin{center}
		\includegraphics[width=\linewidth]{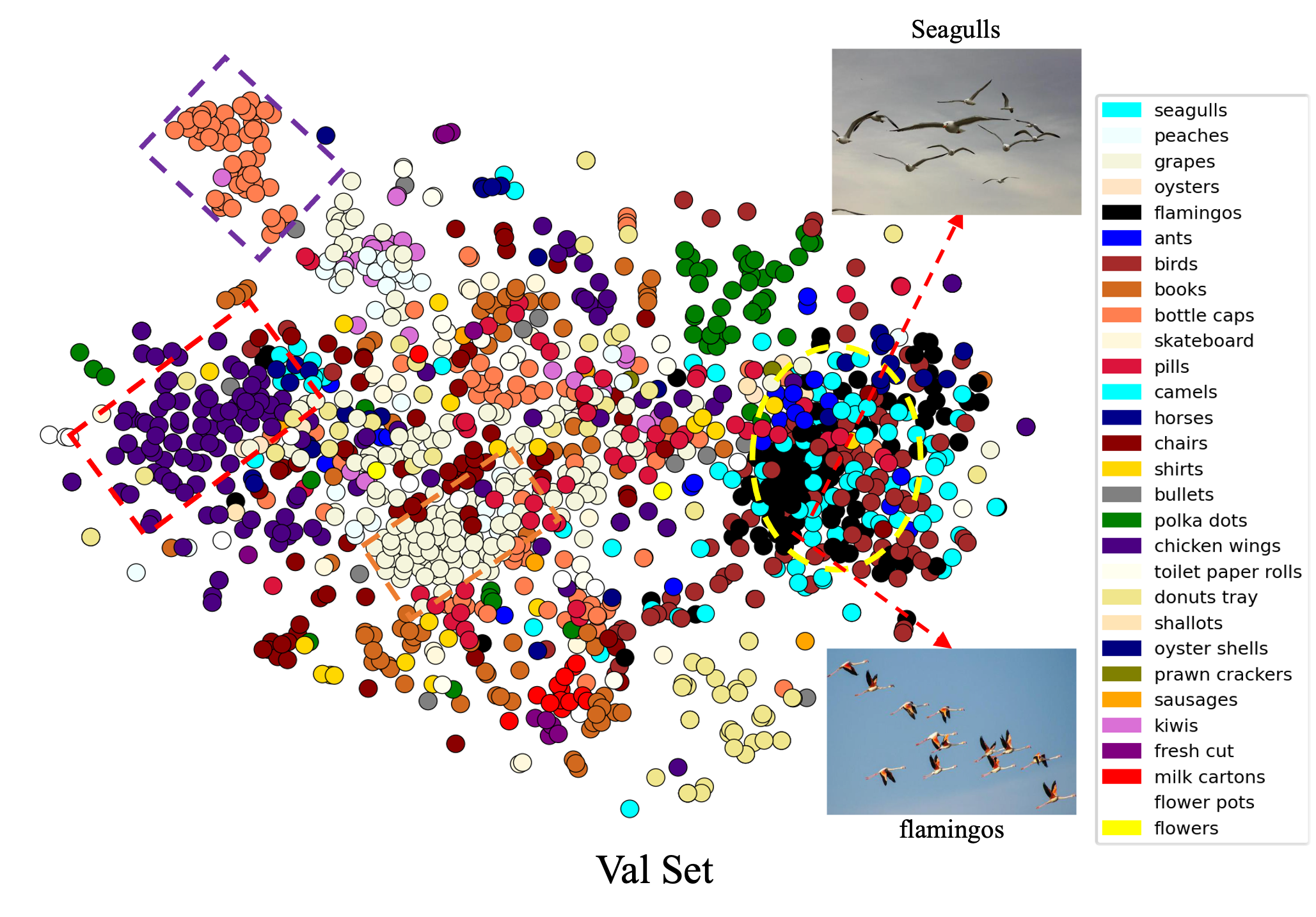}
	\end{center} 
	\caption{Visualization of intermediate recalibration weights learned from dual-attention condenser on the \emph{validation} subset. The result shows that the weights learned for examples of the same classes tend to be clustered together (shown in dashed rectangles).} 
	\label{fig:tsne}
\end{figure}

\begin{figure}[h]
	\begin{center}
		\includegraphics[width=\linewidth]{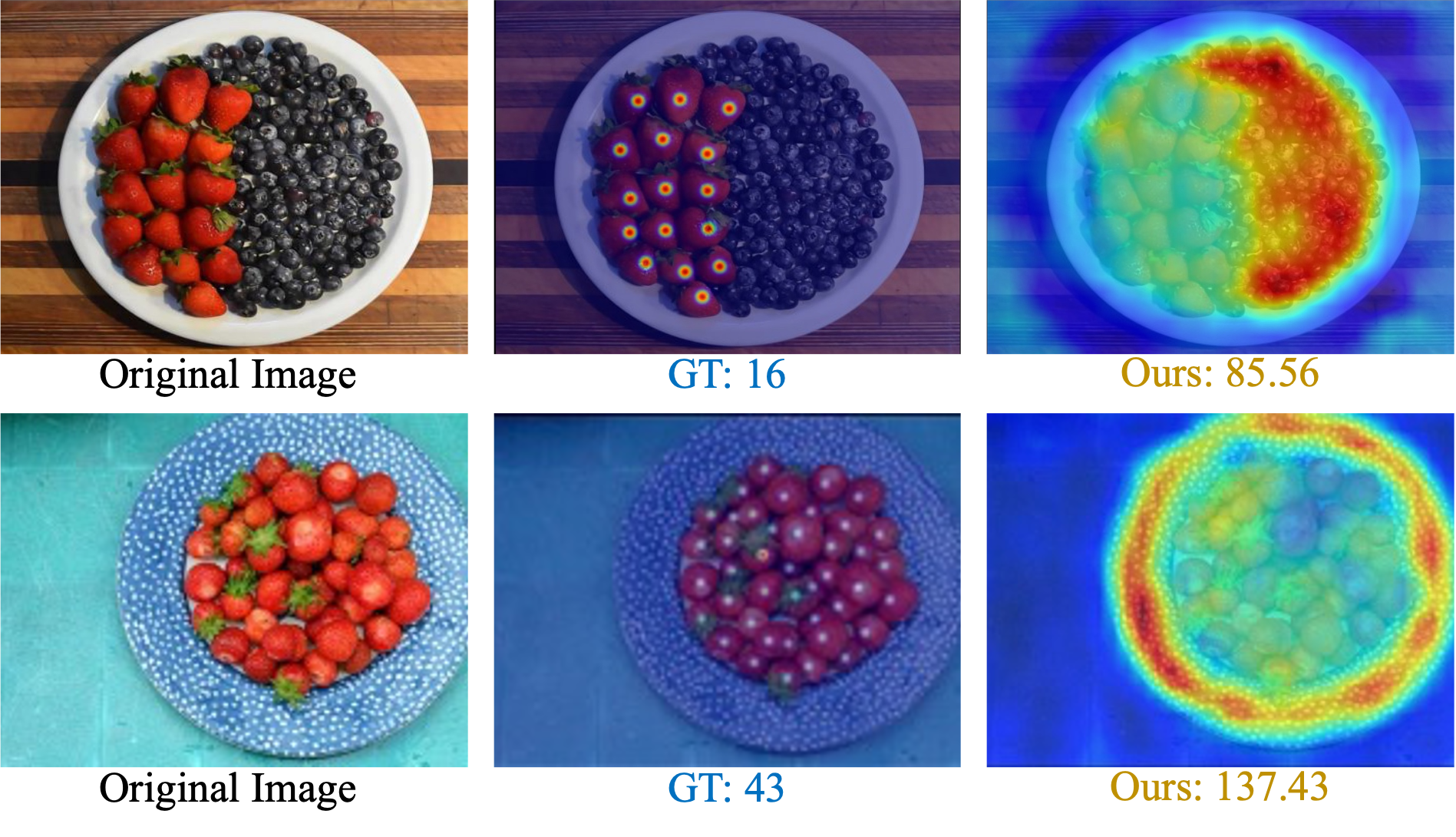}
	\end{center} 
	\caption{Failure cases where GCNet confuses which objects should be counted, \ie strawberries \vs blueberries and strawberries \vs ~decorative dots.} 
	\label{fig:failure}
\end{figure}
To understand the effectiveness of the proposed dual-attention recalibration, we visualize the learned attention weights for recalibrated anisotropic features and pseudo exemplar tokens using T-SNE; see Fig.~\ref{fig:tsne}. The weights learned for the same classes tend to be clustered together even without feeding in category labels. This suggests that our model is capable of capturing inherent similar attributes among instances of the same classes by representing discriminative attention weights for different dimensions. Nevertheless, due to the lack of class annotations, GCNet does not try to distinguish among classes with similar appearances (\eg seagulls \vs flamingos), resulting in these classes being grouped together.
Without user-specified exemplars, GCNet sometimes gets confused with the counting objective as well. As shown in Fig.~\ref{fig:failure}, when the goals are counting strawberries only, GCNet finds self-similarity among blueberries and decorative dots on the edge of a plate as well and outputs inflated counts \wrt the ground truth. 

\section{Conclusion and Future Work}
In this paper, we develop a novel zero-shot CAC counter that exploits self-similarity learning of inherent repetition patterns. GCNet is highly-generalized and widely-applicable since both exemplar-free and weakly-supervised properties are achieved. Additionally, even without location priors, GCNet is spatially-aware and can produce accurate self-similarity maps. Albeit the impressive performance, the GCNet tends to count all repeating instances across the images, \eg strawberries and blueberries on the same plate, since it does not ask users for classes. In the future,  we plan to incorporate existing research on visual grounding, so that users can specify what to count using keywords or points of interest. This could provide additional flexibility to GCNet, without the drawbacks of requiring exemplars as inputs.

\bibliographystyle{IEEEtran}
\bibliography{egbib}

\end{document}